\documentclass[nonatbib]{article}

\usepackage[final]{nips_2016}

\usepackage[utf8]{inputenc} 
\usepackage[T1]{fontenc}    
\usepackage{url}            
\usepackage{booktabs}       
\usepackage{amsfonts}       
\usepackage{nicefrac}       
\usepackage{microtype}      

\usepackage{amsmath}
\usepackage{amsfonts}

\usepackage{graphicx}
\usepackage{wrapfig}

\usepackage[numbers]{natbib}

\graphicspath{{./}}

\usepackage{hyperref}       

\newcommand{\mb}[1]{\mathbf{#1}}

\title{Learning Deep Architectures for Interaction Prediction in Structure-based Virtual Screening}

\author{
  Adam Gonczarek
  , Jakub M. Tomczak, Szymon Zar\k{e}ba, Joanna Kaczmar \\
  Wroc\l aw University of Science and Technology\\
  \texttt{adam.gonczarek@pwr.edu.pl}
  \And
  Piotr D\k{a}browski, Micha\l\ J. Walczak\\
  Indata SA
  }

\begin{document}

\maketitle

\begin{abstract}
We introduce a deep learning architecture for structure-based virtual screening that generates fixed-sized fingerprints of proteins and small molecules by applying learnable atom convolution and softmax operations to each compound separately. These fingerprints are further transformed non-linearly, their inner-product is calculated and used to predict the binding potential. Moreover, we show that widely used benchmark datasets may be insufficient for testing structure-based virtual screening methods that utilize machine learning. Therefore, we introduce a new benchmark dataset, which we constructed based on DUD-E and PDBBind databases.

\end{abstract}


\section{Introduction}

Virtual screening is one of the leading methods in computational drug discovery, which aims at identification of novel small molecules that are capable of binding a drug target, usually a protein. In short, there are two main approaches of virtual screening, ligand-based and structure-based. Ligand-based virtual screening relies on empirically established data, which provide information on active (binding compounds later called ligands) and inactive (not binding) molecules. This approach exploits chemical and spatial similarity among binders to identify new ligands of proteins. The second approach, structure-based virtual screening, requires structural information of a protein to dock a ligand candidate in the binding pockets of a target. Here, a large number of small molecules is screened against a structure of a target protein. Then, binding capacity between protein and compounds is assessed using scoring functions, and finally compounds are triaged according to their binding potential. 

The main hurdles affecting virtual screening is complexity of chemical space comprising up to $10^{60}$ theoretical \citep{BMG:96} and $10^7$ of commercially available compounds \citep{ISMBC:12}\footnote{\url{http://zinc15.docking.org/}}, as well as high false positive rate of identified ligands and a lack of exhaustive training datasets.

Although the above mentioned hindrances are tackled by various approaches, \textit{e.g.} Smina \citep{KBC:13}, with different success rate, it is the advent of deep learning that promises superior performance in high-throughput virtual screening \citep{UetAl:14}. Deep learning has already been successfully employed in ligand-based virtual screening \citep{DMIBHAA:15, MSLDS:15, RKRWKP:15} but only recently the very first attempts to the structure-based methods have emerged \citep{PCS:16, WDH:15}. 

In this study, we propose \textbf{a new deep architecture for predicting binding capacity of a protein-molecule  pair}. In addition, we demonstrate the disadvantages of common benchmark datasets, which are used for training and testing screening methods. To fill this gap, we propose a \textbf{new benchmark dataset} that is more suitable for structure-based virtual screening.

\section{Methodology}

\begin{figure}[!hpbt]
\centering
\includegraphics[width=0.8\textwidth]{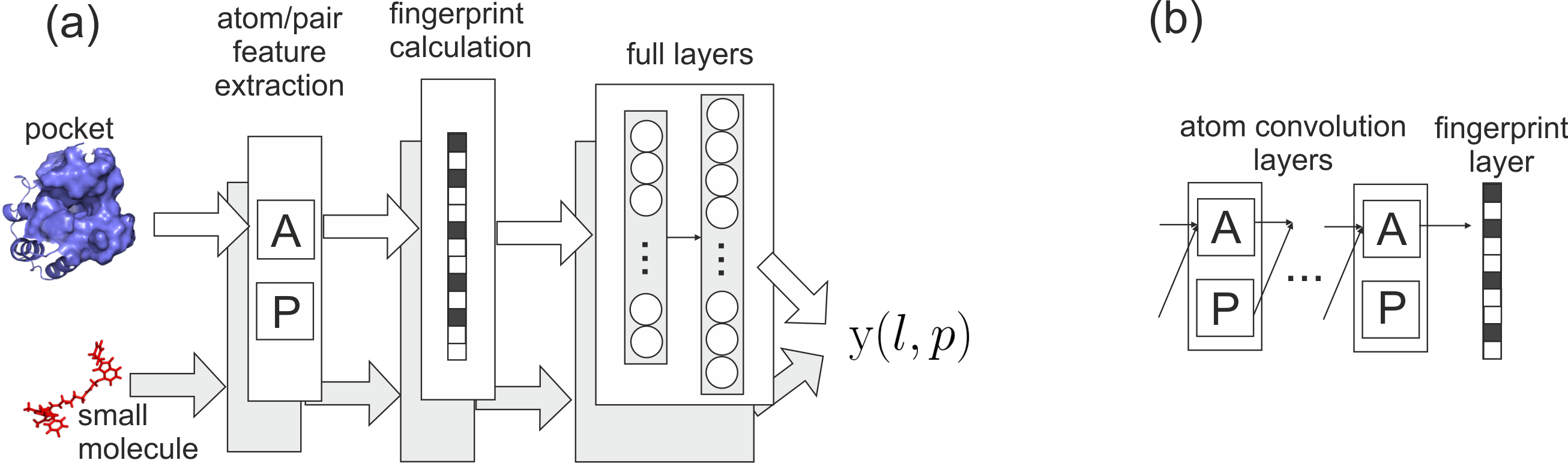}
\caption{(a) Schema of the proposed approach. Letters A and P denote lists of atoms and connections, respectively. (b) Details about the neural fingerprint.}
\label{fig:pipeline}
\end{figure}

\paragraph{Model} Our aim is to predict binding potential $y \in \{0,1\}$ for given pair of a small molecule $l$ and a target (represented by a pocket in protein structure to which a ligand may bind) $p$. We face three major problems in the stated task: \textbf{(i)} both target and small compound vary in size, \textbf{(ii)} each of them is represented by a list of atoms and therefore a method must be invariant to any permutation of the list, \textbf{(iii)} these are 3D structures, thus a method must be invariant to translations and rotations. To cope with these issues we propose to process the protein and the small molecule separately to obtain two fixed-size descriptions (\textit{fingerprints}) that can be further transformed non-linearly \textit{e.g.} by neural nets, and finally used for binding prediction. This approach aims at processing the protein-compound pair separately and then learning a relation of the interaction. A pipeline of the proposed method is presented in Figure \ref{fig:pipeline}(a).

The crucial part of the proposed approach is a \textit{fingerprint}, \textit{i.e.}, a description of a fixed size. One of the widely used fingerprints for virtual screening is \textit{Extended Connectivity Fingerprint} (ECFP) \citep{RH:10}. ECFP is an automatic manner of determining fingerprints by consecutively applying a \textit{hash function} on atom and its neighborhood followed by a \textit{indexing operation}. The hash function allows to combine information about each atom and its neighboring substructures while the indexing operation is used to combine all the nodes' features into a single fingerprint of the whole compound. However, due to pre-determined form of hashing and indexing, ECFP is sensitive to small perturbations in molecule structure, and therefore the features obtained by this method are not very robust.

Very recently, the drawbacks of ECFP were alleviated by application of learnable operations similar to operations in convolutional neural nets \citep{DMIBHAA:15}. Here the hashing is replaced with an adaptive convolutional-like operation and the indexing with a softmax operation. This could be formalized as follows. Let us denote an $m^{th}$ atom in a compound described by $F$ features by $\mb{a}_{m}$. Then hashing could be described in the following fashion:
\begin{equation}\label{eq:atom_convolution}
\mb{a}_{m} := \sigma\left( \mb{W} \mb{a}_{m} + \sum_{i=1}^{I_{m}} \mb{H}_{I_{m}}\mb{a}_{i} + \mb{b} \right) ,
\end{equation} 
where $\mb{a}_{i}$ is $i^{th}$ neighboring atom, $I_{m}$ is the number of possible neighbors for the $m^{th}$ atom\footnote{Due to the physical properties of compounds there can be only up to $5$ neighbors, so $I_{m} \in \{1, 2, \ldots, 5\}$.}, $\mb{W} \in \mathbb{R}^{R \times F}$ is a matrix of weights\footnote{Notice that $R$ is the number of new features and this could differ from $F$.}, $\mb{H}_{1}, \ldots, \mb{H}_{5} \in \mathbb{R}^{R \times F}$ are matrices of weights for neighbors, $\mb{b} \in \mathbb{R}^{R}$ is a bias vector, and $\sigma(\cdot)$ is an element-wise non-linear function, \textit{e.g.}, the sigmoid function or ReLU. We refer to this operation as \textit{atom convolution} and it can be repeated $K$ times which constitutes $K$ layers, and each layer consists of own weights to learn (see Fig. \ref{fig:pipeline}(b)).

The indexing operation is then replaced with a softmax operation that consecutively applies the softmax function to each atom in the compound to yield the final \textit{neural fingerprint} $\mb{n}$:
\begin{equation}\label{eq:neural_fingerprint}
\mb{n} = \sum_{m} \mathrm{sofmax}(\mb{V} \mb{a}_{m} + \mb{c})
\end{equation}
where $\mb{V} \in \mathbb{R}^{S \times R}$ is a weight matrix, $\mb{c} \in \mathbb{R}^{S}$ is a bias vector and $S$ is the size of the fingerprint.

Next, after obtaining the neural fingerprint for small compound ($\mb{n}_{l}$) and protein ($\mb{n}_{p}$), we apply a neural network (MLP) to obtain new representations: $\mathbf{w} = \mathrm{MLP}_{l}(\mb{n}_{l})$ and $\mathbf{v} = \mathrm{MLP}_{p}(\mb{n}_{p})$. Eventually, we calculate the bioactivity by transforming the inner product of $\mb{w}$ and $\mb{v}$ using the sigmoid function $\mathrm{sigm}(\cdot)$:
\begin{equation}\label{eq:prediction}
y(l, p) = \mathrm{sigm}( \mb{w}^{\top} \mb{v} ).
\end{equation}

\paragraph{Training} Standard learning of neural networks utilizes the cross-entropy (CE) loss function. Typically databases contain only active ligand-protein pairs. Hence, we propose to add an additional term to the CE loss in order to avoid overfitting to the positive class:
\begin{equation}\label{eq:nc_loss}
\mathcal{L}(\theta) = \frac{1}{N} \sum_{n=1}^{N} \log y(l_{n}, p_{n}) + \mathbb{E}_{l,p \sim P(l,p)} [ \log (1 - y(l, p)) ].
\end{equation}
Despite the fact that the expected value can be approximated using Monte Carlo methods, this formulation causes a problem since finding the joint probability of the small molecule-protein pair is a very complex task. We overcome this issue with the assumption that taking a random pair from the dataset would result in a negative (not binding) example. Obviously, such an approach introduces a bias, however, a chance of producing wrong label is negligible. The proposed loss function in (\ref{eq:nc_loss}) is closely related to the Noise Contrastive Estimation \citep{GH:12}.

\section{Results}

The efficacy of machine learning methods for virtual screening are typically evaluated with one of the renowned benchmarks, \textit{e.g.}, DUD-E \citep{MCIS:2012}. Generally, the evaluation dataset is divided into training and testing sets that contain different targets (together with their actives and decoys). Interestingly, it turns out that this testing protocol might be strongly biased due to similarity of artificial decoys for different targets. We addressed this problem with a newly developed benchmark based on two separate datasets.

\begin{wraptable}{r}{0.5\textwidth}
  \caption{Results on DUD-E benchmark  ($70\%$ of data for training and $30\%$ of data for testing) and on DUD benchmark (leave-one-out cross-validation).}
  \label{tab:old_benchmark}
  \centering
  \begin{tabular}{ccc}
    Dataset & Method     	& Mean AUC\\
    \hline
    DUD-E & Smina 		& $0.700$	\\
   & AtomNet \citep{WDH:15} 	& $0.855$ \\
    & cmpds ECFP + LR & \textbf{0.904} \\
    \hline
    DUD & DeepVS \citep{PCS:16} 	& $0.800$	\\
  \end{tabular}
\end{wraptable}

\paragraph{DUD-E experiment} We used DUD-E\footnote{\url{http://dude.docking.org/}} benchmark, consisting of $102$ proteins (targets), $22,886$ active compounds (ligands or binders) and over $1M$ decoys (non-binders). We randomly divided targets into training ($72$) and testing ($30$) parts, which is similarly to \citep{WDH:15}. We applied the ECFP fingerprint with the size of $4096$ to small compounds (cmpds) only and trained logistic regression (LR) to discriminate between actives and decoys. Notice that no information about targets was used. The method achieved $0.904$ mean AUC, evaluated on the targets in the test set, and to the best of our knowledge it has outperformed other state-of-the-art methods for structure-based virtual screening trained in the similar manner (see Table \ref{tab:old_benchmark}). Thus, this suggests that datasets with many artificially generated decoys (like DUD-E) are prone to bias due to similarity of majority of the inactive compounds for one target to inactive compounds for other targets. Further, application of basic learning methods to small compounds only results in improved performance. Consequently, it is uncertain whether a method evaluated on this testing scheme learns the relationship between compounds and targets, or learns the discrimination between active and inactive molecules, where additional information about targets only contributes to noise.

\paragraph{PDBBind + DUD-E} Next, we employed PDBBind \citep{LLHLLZNLW:2014} for training and DUD-E for testing. The original PDBBind database contains 3D structures of about $10k$ complexes, \textit{i.e.}, structures of ligands docked in binding pockets of proteins. We have removed all the complexes containing targets from DUD-E, obtaining $8822$ complexes for training. Notice that this dataset contains no artificially generated decoys, negative examples are generated by sampling random target-compounds pairs from the dataset (see Eq. \ref{eq:nc_loss}). For testing we used $88$ DUD-E targets represented by a binding pocket extracted from the original PDBBind. For each target we have randomly sampled $1000$ compounds (actives and decoys) from DUD-E, resulting in $88,000$ testing examples. We tested two different models based on the pipeline presented in Fig. \ref{fig:pipeline}(a). In the first model, standard ECFP fingerprints were applied both to small compounds and pockets. In the second one we adopted learnable neural fingerprints. We compared our approach to two widely used methods, \textit{i.e.}, AutoDock Vina \citep{TO:2010} and Smina \citep{KBC:13}. The results are presented in Table \ref{tab:new_benchmark}. First, we see the importance of applying learnable fingerprint, since ECFP performed worse than the reference methods. Second, we observe that the neural fingerprint-based approach outperformed both reference methods, achieving better mean AUC, and reaching more targets that exceeded high AUC thresholds.           

\begin{table}[!hpbt]
  \caption{Results obtained on the proposed benchmark. The presented approach with ECFP is denoted by Ours(ECFP) and the one with neural fingerprint by Ours(NF). AUC $\geq \alpha$ denotes on how many targets (out of $88$) a method performs better than $\alpha$.}
  \label{tab:new_benchmark}
  \centering
  \begin{tabular}{cccccc}
    Method     	& Total AUC & Mean AUC ($\pm$ std.) & AUC $\geq 0.7$ & AUC $\geq 0.8$ & AUC $\geq 0.9$\\
    \hline
    AutoDock Vina & $0.644$  	& $0.691 \pm 0.147$ &	47 & 21 & 4\\
    Smina 		& $0.653$  	& $0.704 \pm 0.138$ & \textbf{54} & 23 & 4\\
    Ours(ECFP) 	& $0.600$ 		& $0.551 \pm 0.166$  & 21 & 2 & 0\\
    Ours(NF) 	& \textbf{0.714} 		& \textbf{0.705} $\pm$ \textbf{0.168}  & 47 & \textbf{29} & \textbf{11}\\
  \end{tabular}
\end{table}

\section{Conclusions}

This study results in two contributions in the field of computational drug discovery. First, we propose a new benchmark dataset built on top of the two established datasets, PDBBind and DUD-E. This benchmark provides suitable information for the development of structure-based virtual screening methods. At the same time, we demonstrate that currently available benchmarks represent mediocre training and testing sets due to insufficient coverage of chemical complexity. Second, we propose a novel deep learning-based approach able to identify ligands of target protein. The performed experiments showed that our approach outperforms two widely used methods, AutoDock Vina and Smina. Here developed method reaches AUC of $0.9$ or greater for the $11$ targets, while the reference methods exceed AUC of $0.9$ for the $4$ targets. We anticipate further evolution of the proposed approach by applying more sophisticated deep learning techniques, \textit{e.g.} by developing more accurate learnable fingerprints. 

 \subsubsection*{Acknowledgments}
The work conducted in this paper is partially co-financed by European Regional Development Fund within the framework of the Smart Growth Operational Programme 2014-2020, grant No. POIR.01.01.01-00-1083/15, and by European Union FI-Core project, grant no. 632893, (FIWARE@Wroclaw), funded within FI-PPP and FP7.

%
%
\bibliographystyle{plainnat}
\bibliography{nips2016}

\end{document}